\documentclass{article}

    \usepackage[final, nonatbib]{neurips_2020}

\usepackage[utf8]{inputenc} 
\usepackage[T1]{fontenc}    
\usepackage{hyperref}       
\usepackage{url}            
\usepackage{booktabs}       
\usepackage{amsfonts}       
\usepackage{nicefrac}       
\usepackage{microtype}      
\usepackage{multirow}
\usepackage{makecell}
\usepackage{graphicx}
\usepackage{chngcntr}
\usepackage{hyperref}

\title{LandCoverNet: A global benchmark land cover classification training dataset}

\author{%
  Hamed Alemohammad \\
  Radiant Earth Foundation\\
  \texttt{hamed@radiant.earth} \\
   \And
  Kevin Booth \\
  Radiant Earth Foundation\\
  \texttt{kevin@radiant.earth} \\

}

\begin{document}

\maketitle

\begin{abstract}
Regularly updated and accurate land cover maps are essential for monitoring 14 of the 17 Sustainable Development Goals. Multispectral satellite imagery provide high-quality and valuable information at global scale that can be used to develop land cover classification models. However, such a global application requires a geographically diverse training dataset. Here, we present LandCoverNet, a global training dataset for land cover classification based on Sentinel-2 observations at 10m spatial resolution. Land cover class labels are defined based on annual time-series of Sentinel-2, and verified by consensus among three human annotators. 

\end{abstract}

\section{Introduction}

Land cover (LC) maps play an important role in many applications such as precision agriculture, urban mapping, and environmental monitoring among others~\cite{Gomez2016}. Earth observations provide valuable data at global scale that can be used to build applications for LC mapping. Recently, with the advancements in machine learning (ML) techniques, several studies have been carried out to develop ML-based LC classification models~\cite{Pelletier2019, Fu2017, Robinson_2019_CVPR, Carranza_Garc_a_2019}. However, the remaining challenge is lack of high-quality open-access training data that represents the geographical diversity of LC classes at global scale~\cite{shankar2017classification, Gong2016}. Such a training dataset can open the value of open satellite imagery (such as Landsat and Sentinel missions) for regional and global monitoring with enhanced accuracy. 

In response to this need, we present LandCoverNet the first open-access, global and geographically diverse LC classification training dataset. LandCoverNet is based on multi-spectral imagery from Sentinel-2 satellites, and consists of globally representative image chips and labels. Specification of the dataset, including LC taxonomy, is defined from the outputs of an expert workshop that we hosted two-years ago.

Generating labels for a large-scale training dataset such as LandCoverNet requires extensive coordination with annotators, and deploying tools to facilitate pixel-level labeling. Moreover, human interpretation error is unavoidable at 10 m spatial resolution. Therefore, a consensus algorithm was implemented to generate the label for each pixel in the dataset using inputs from multiple users. 

In this paper, we discuss the label generation procedure, the sampling scheme, and specifications of the data. LandCoverNet v1.0 which covers the whole African continent is now available for public access, and other parts of the world will be gradually released.

\section{Dataset Specifications}

\subsection{Land Cover Taxonomy}
LC class taxonomies vary across existing global LC products. While there are multiple taxonomies for LC classification, they do not take into account the spatial resolution at which the classification map will be produced. This is essential since some classes might not be resolvable with the resolution of the input satellite imagery. 

To address this problem, we organized an expert workshop to define the specification of LandCoverNet dataset as a benchmark for the community. The participants who were domain experts in LC mapping and ML developed a hierarchical LC class taxonomy. Based on the hierarchical definitions, they recommended the taxonomy presented in Table~\ref{taxonomy} for LandCoverNet dataset. 

Class definition in the taxonomy presented in Table~\ref{taxonomy} is based on annual time-series of each location. This means in order to assign one of these classes to a region, one needs to examine the annual time-series of observations from the satellite, and then assign the appropriate class. 

\begin{table}
  \caption{LandCoverNet class taxonomy}
  \label{taxonomy}
  \centering
  \begin{tabular}{lll}
    \toprule
    Level 1     & Level 2     & Level 3 \\
    \midrule
    \multirow{4}{*}{\makecell{Bare \\ max veg/yr $\le 10\%$}} & Snow/Ice  & Snow/Ice     \\
    \cmidrule(r){2-3}
     & Water & Water\\
     \cmidrule(r){2-3}
     & \multirow{2}{*}{Bare Ground} & Artificial \\
     \cmidrule(r){3-3}
      & & Natural \\
    \midrule
    \multirow{4}{*}{\makecell{Vegetation \\ max veg/yr $\geq$ 10\%}} & Woody & Woody \\
    \cmidrule(r){2-3}
    & \multirow{2}{*}{Non-Woody} & Cultivated \\
    \cmidrule(r){3-3}
    &  & (Semi) Natural \\
    \bottomrule
  \end{tabular}
\end{table}

\subsection{Sentinel-2 Tile Sampling}
Sentinel-2 data are provided as $100 \times 100$ $km^2$ ortho-images in UTM/WGS84 projection. Each of the $100 \times 100$ $km^2$ regions is called a tile. To build a geographically diverse training dataset, we designed a sampling scheme and selected 300 tiles of Sentinel-2 for training data generation. 

The number of samples in each continent is set to be proportional to the area of the continent. To capture diversity of land surface in each continent, we used an existing global LC map at 250m resolution based on MODIS instrument~\cite{Sulla-Menashe2019} to define LC classes within each tile of Sentinel-2. We chose to use MODIS LC maps as features for sampling our regions as this is the only way to ensure diversity in our samples. Practically, MODIS LC data is the best estimate of the distribution of LC classes at global scale, and we use it to sample a set of tiles and generate an accurate training dataset that can be used for future LC mapping. 

For each tile of Sentinel-2 (approximately 18000 across all land areas), we calculated a set of features as following: latitude and longitude of the tile centroid, and percentage of each of the 17 IGBP LC classes in the tile based on MODIS product. This results in 19 features for each tile. To ensure tiles do not have a dominant LC class, any tile with more than 60\% water or 90\% barren land was excluded from the sampling. 
Finally, we used the empirical CDF of the features, and uniformly sampled the tiles in each continent. The resulting data contains 66 tiles in Africa, 92 tiles in Asia, 20 tiles in Australia and Oceania, 28 tiles in Europe, 54 tiles in North America and 40 tiles in South America. 

\subsection{Chip Selection}
As part of the expert workshop, and a follow up user survey, the chip size for this training dataset was set to $256 \times 256$. For the first set of these labels, and based on computational resources we decided to generate 30 chips within each of the 300 selected tiles of Sentinel-2. This results in 9000 chips at global scale, or equivalently $\sim$589 million pixels. 

\section{Methodology}
Generating human verified labels for time-series imagery is a challenging task. In order to facilitate the process, we provided our team of annotators with a "guess label" based on a model prediction. For this, we trained a Random Forest model for each tile of Sentinel-2 in our dataset, and overlaid that in the annotation dashboard. Details of this model training is explained in Appendix~\ref{modeling}. In this section, we explain the human validation of labels and calculation of the consensus score. 

\subsection{Human Validation}
To generate a high-quality training dataset, we validated labels of each pixel with the help of human annotators. For this, we worked with a team of annotators who had experience labeling satellite imagery, and provided them with training necessary to work on this dataset, and our labeling dashboard. We also performed a labeling campaign with them on a test dataset, to ensure they are familiar with the taxonomy before starting on the main dataset. The new element for this labeling task was the time-series dimension. 

Every user was asked to look through all the 24 scenes of Sentinel-2 for each chip and then check the model prediction. If the predictions are good, leave them unchanged, otherwise add the correct label. They had to submit the labels even if no changes were made to ensure the label was marked as validated. 

\begin{figure}
  \centering
  \includegraphics[width=0.8\linewidth]{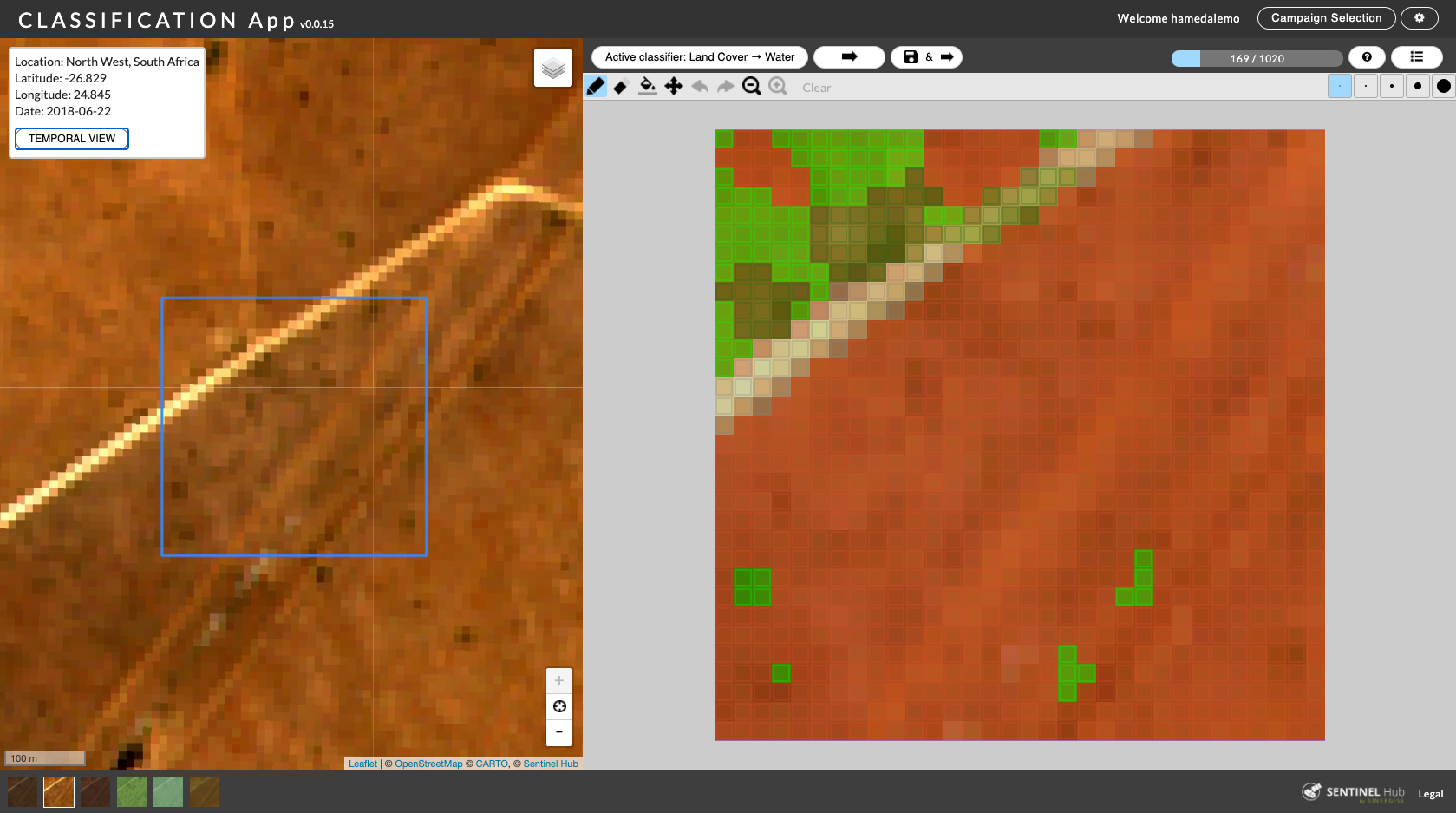}
  \caption{Labeling dashboard showing Sentinel-2 scene on the left side and the labels on the right.}
  \label{dashboard_1}
\end{figure}

\begin{figure}
  \centering
  \includegraphics[width=0.8\linewidth]{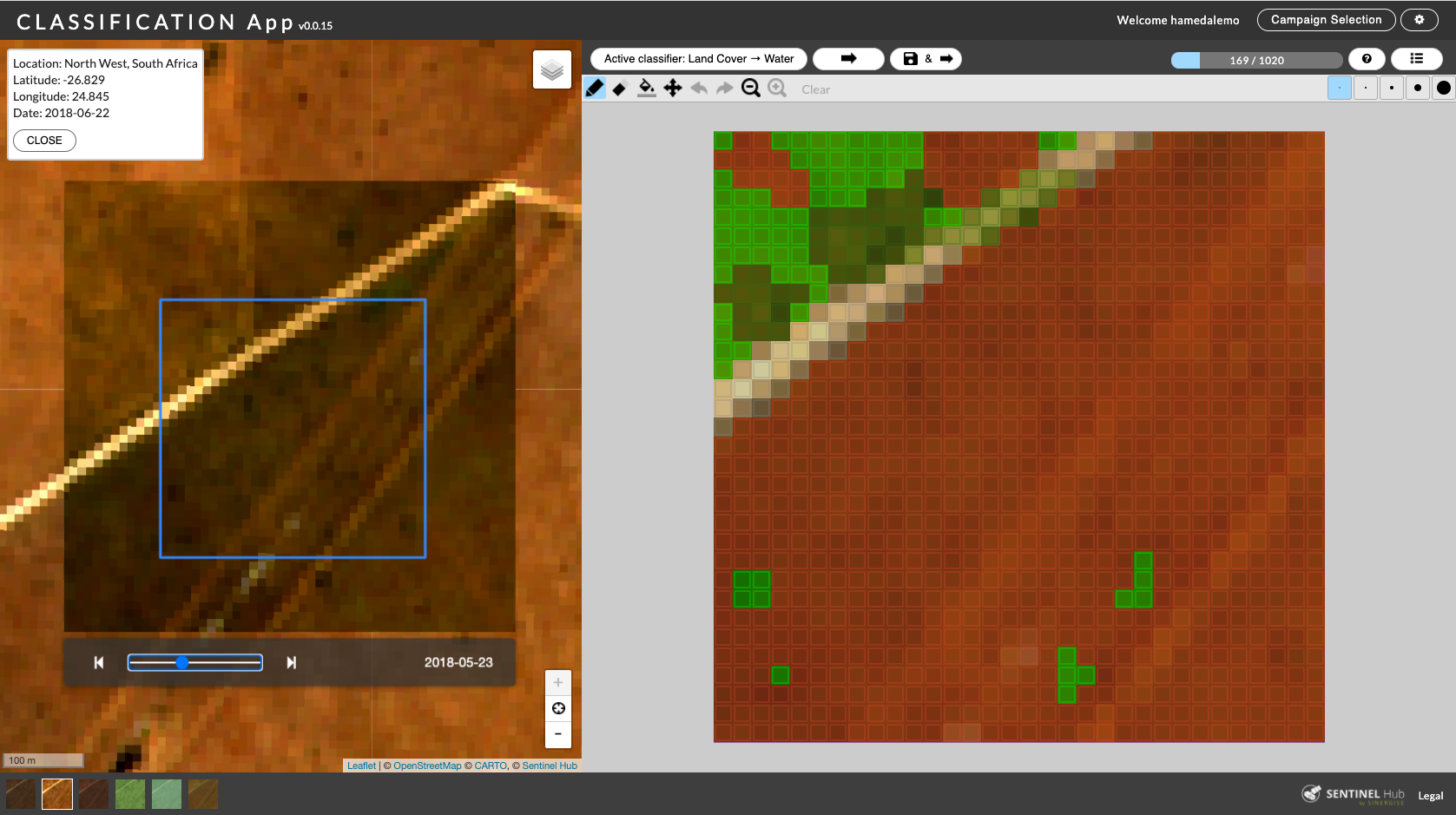}
  \caption{Same as Figure~\ref{dashboard_1} with the temporal window overlaid on Sentinel-2 scene allowing users to select different times of the year.}
  \label{dashboard_2}
\end{figure}

\begin{figure}
  \centering
  \includegraphics[width=0.8\linewidth]{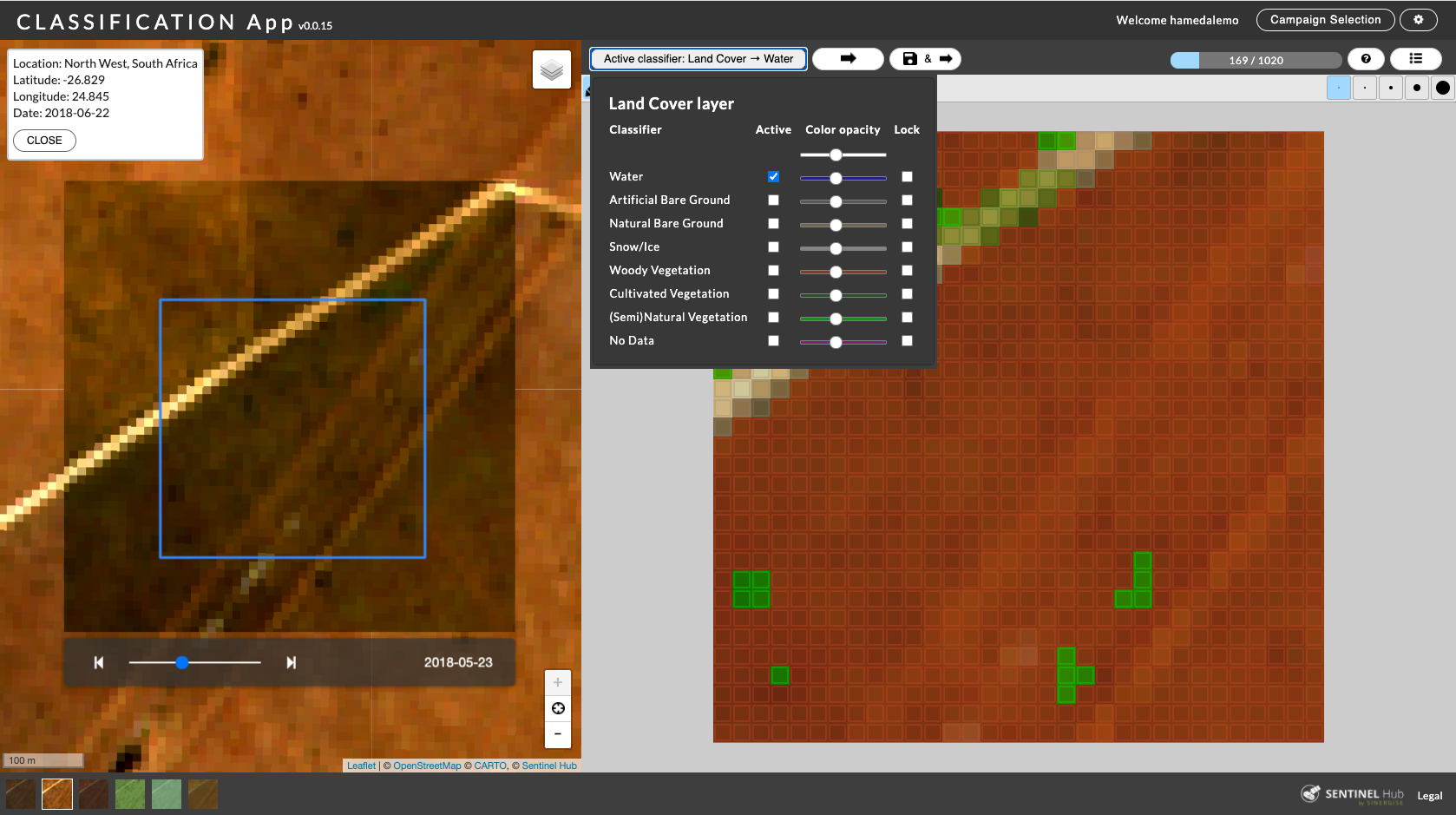}
  \caption{Same as Figure~\ref{dashboard_2} with the class selection tool open to show how users select different classes.}
  \label{dashboard_3}
\end{figure}

To facilitate the labeling, each $256 \times 256$ chips was split to 64 tasks of $32 \times 32$ pixels. In each task, annotators would see the Sentinel-2 RGB image, false color NIR visualization, NDVI and NDWI as well SWIR bands. They had the option to zoom in and out of the task bounding box and examine the surrounding area to help them with the labeling. Figures~\ref{dashboard_1} - \ref{dashboard_3} show screenshots of the labeling dashboard.  

We implemented another measure to minimize chances of error in human validation of our data~\cite{Elmes_2020}. Each chip was sent to three individuals in the annotation team and they were asked to validate the labels. Then we used a consensus technique (Section~\ref{consensus}) to generate the final label for each pixel. 

\subsection{Consensus Label}
\label{consensus}
Sentinel-2 imagery has a 10m spatial resolution in the RGB and Near Infrared (NIR) bands. While distinguishing between the 7 classed outlined in Table~\ref{taxonomy} is possible at this resolution, human error is inevitable. As discussed by Elmes et al.~\cite{Elmes_2020}, such errors need to be taken into account during the training data generation process, and quantified in the final product. 

To mitigate these error, we took two steps in our label validation phase. First, about 540 tasks were also labeled by an expert in our team to be used as reference for assessing each annotators' work. Second, each task was independently labeled by three annotators.

To derive a consensus score and label for each pixel using the three labels, we first calculated a score for each annotator using the expert labeled tasks. Annotator's score for each task is defined as the percentage of pixels that they correctly labeled in reference to the expert labels. Using the 540 expert labeled tasks, we calculated an overall accuracy score for each annotator. 

Finally, using a Bayesian averaging approach we calculate the consensus probability for each of the 7 classes. Then the class with highest probability is selected as the label and the probability is reported as the consensus score~\cite{Estes2016, Debats2016}. Eq.~\ref{eq:score} presents the consensus score for a pixel:

\begin{equation}
    P(\theta_j) = \sum_{w_i=1}^{n} P(w_i) \times P(\theta_j|w_i)
    \label{eq:score}
\end{equation}
where:
$\theta_j \in [1, \dots 7]$ is the true LC class, $n$ is number of annotators labeling a pixel, and $P(w_i)$ is the accuracy score of annotator $w_i$. 

\section{Dataset Properties}
\begin{figure}
  \centering
  \includegraphics[width=0.7\linewidth]{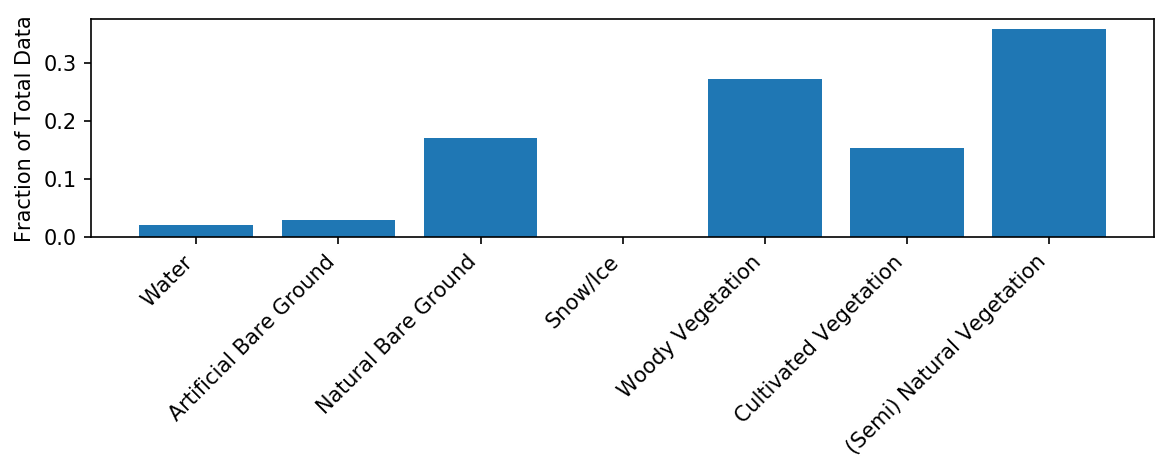}
  \caption{Distribution of different classes in the LandCoverNet v1.0}
  \label{label_dist}
\end{figure}

\begin{figure}[ht]
  \centering
  \includegraphics[width=0.5\linewidth]{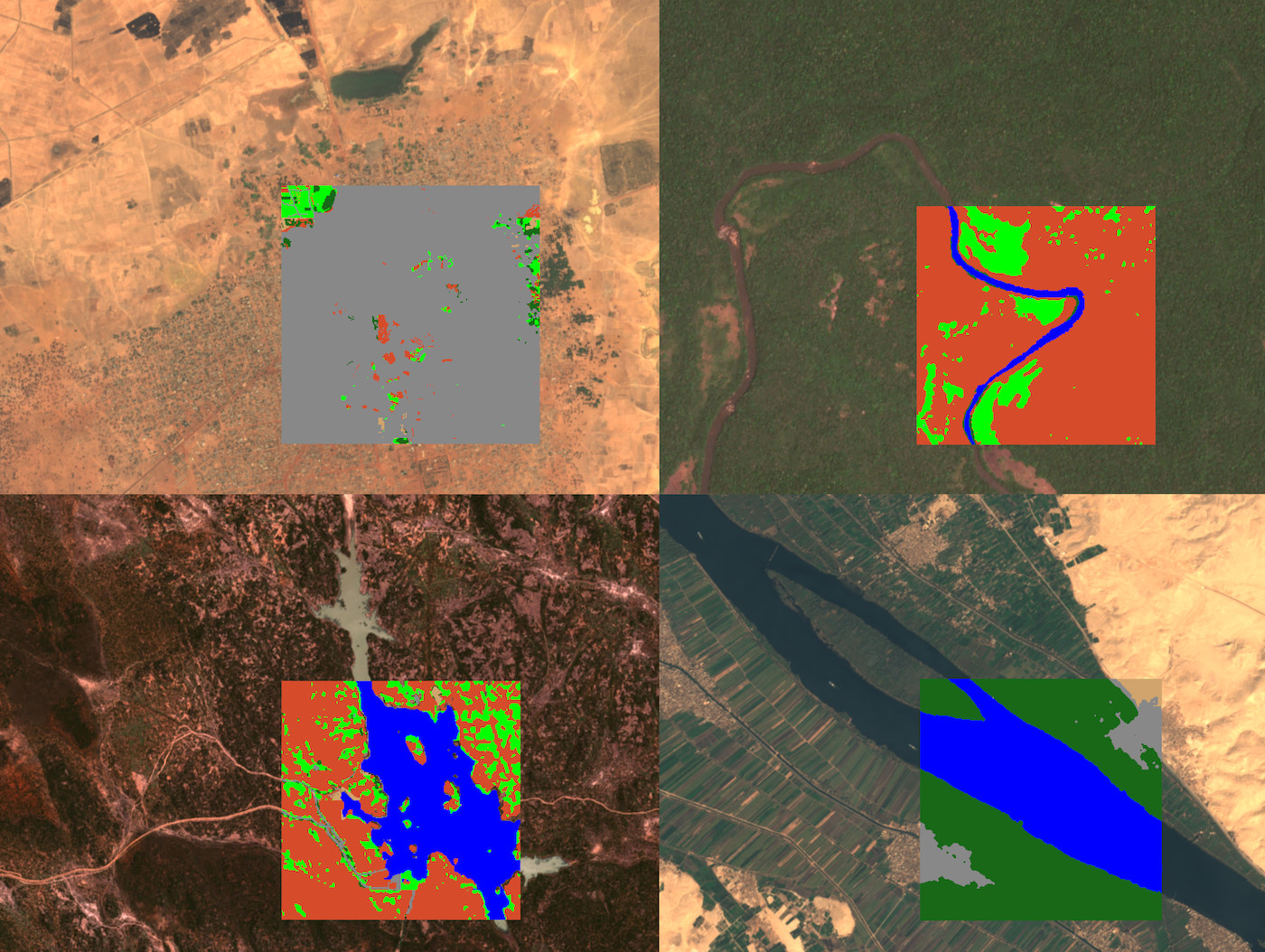}
  \caption{Four sample chips of LandCoverNet labeled data (color shaded regions) overlaid on one of their Sentinel-2 scene. (blue: water, dark green: cultivated vegetation, light green: (semi) natural vegetation, grey: artificial bare ground, brown: woody vegetation)}
  \label{samples}
\end{figure}

LandCoverNet v1.0 which covers all the image chips for Africa is published on Radiant MLHub for public access. This version includes 1980 chips of $256 \times 256$ pixels across Africa. The labels in the dataset cover all the classes other than permanent snow/ice which is not common in Africa. Figure~\ref{label_dist} shows the distribution of pixels within different classes. Majority of the pixels fall within vegetated class, including about 15\% cultivated pixels. Natural bare ground also has a significant portion with about 17\% of the data. Figure~\ref{samples} shows four sample chips overlaid on Sentinel-2 imagery from the dataset. 

\begin{figure}
  \centering
  \includegraphics[width=0.7\linewidth]{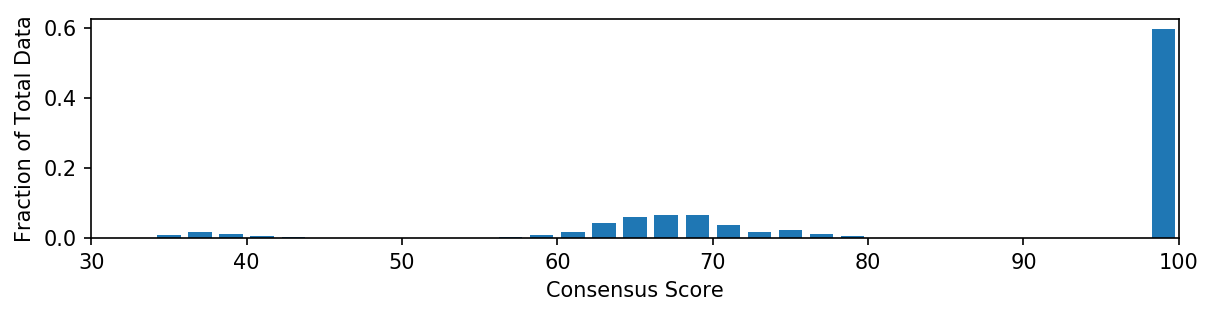}
  \caption{Distribution of consensus score in the LandCoverNet v1.0}
  \label{cons_dist}
\end{figure}

Overall, the consensus scores are very high. Figure~\ref{cons_dist} shows the histogram of consensus scores across all pixels. 60\% of the data has a score of 100\% meaning all annotators agreed on the class, and the rest have relatively high score, with about 34\% of the data having a consensus score of more than 60\% and less than 100\%. Users who access the data get a per pixel consensus score that can be used during model training to increase the model confidence in high score pixels and vice-versa.

\section{Dataset Access}
 LandCoverNet is distributed under a Creative Commons Attribution 4.0 International (CC BY 4.0) license. Interested users can access the data on Radiant MLHub\footnote{www.mlhub.earth} through an API. The dataset has a description page accessible on \url{www.landcover.net} which includes a Jupyter Notebook tutorial for downloading the data.

\section*{Acknowledgements}
This dataset was generated with a grant awarded to Radiant Earth Foundation from Schmidt Futures. We would like to acknowledge in kind technology support from our collaborators at Sinergise.

\bibliographystyle{unsrt}
\bibliography{landcovernet}

\appendix
\counterwithin{figure}{section}
\counterwithin{table}{section}

\section{Generating Guess Labels for Human Validation}
\label{modeling}
\subsection{Model Development}
The goal of this LC classification model was to provide a baseline for annotators to facilitate their labeling. Therefore, we chose a simple Random Forest architecture for this and trained a pixel-wise classification model that used time-series of observation from Sentinel-2 and predicted the LC class. 

\subsubsection{Input Data}
For each tile, we queried the Sentinel-2 catalog in calendar year 2018, and selected 24 scenes for modeling. Scenes were selected in a two step approach: 1) one scene was selected in each calendar month with the minimum cloud cover percentage, and 2) 12 more scenes were selected throughout the year sorted by their cloud cover percentage. The logic for this selection was to ensure there is enough data distributed throughout the year to reconstruct the time-series for each pixel.

We used 10 bands from Sentinel-2 observations as predictors. Red, Green, Blue and NIR bands that are at 10m spatial resolution and Red-edge bands 1-4, and SWIR bands 1-2 that are at 20m spatial resolution. We used a nearest neighbor interpolation to map 20m bands to 10m for each scene.

Sentinel-2 L2A product has a scene classification layer (SCL) that provides a quality flag for each pixel, and a four class label (water, vegetation, snow, not-vegetated). We used this layer as a filter to clean the time-series data before inputting to the model. Any pixel that was classified as cloudy, cloud shadow or saturated/defective was removed from the time-series. 

Finally, all the valid observations of each pixel were used to interpolate the missing or filtered-out ones, and reconstruct a time-series of 24 observations with 10 bands at each time for all pixels within the scene. We used a linear interpolation with a periodicity of 1 year to reconstruct the time-series. These 240 features were used as input in the model training. 

\subsubsection{Model Training}
For training a LC classification model, we needed yet another target LC classification product. Therefore, we decided to use the best available product. Noting that any LC map has its own uncertainties, and our goal was to generate the best guess for our annotators. We selected GlobeLand30 (GLC) product which was produced for the year 2010 using Landsat 8 data at 30m spatial resolution~\cite{JOKARARSANJANI201625}. 

We used the method presented by~\cite{Nachmany2018} to map GLC classes to LandCoverNet taxonomy. To remove any discrepancy between GLC product from 2010 and our imagery from 2018, we used a majority voting on the SCL labels from Sentinel-2 L2A and defined a set of annual 4-class filter for each pixel: water, vegetation, snow, not-vegetated. Comparing these classes to the LC class from GLC for each pixel, we dropped any pixel that had a mis-match between the two labels, and didn't used them in the training. The remaining pixels were split to train (75\%) and test (25\%) sets using stratified sampling. 

The training (and test) sets naturally have class imbalance as the spatial coverage of a tile is limited to $100 \time 100$ $km^2$. Therefore, we used a Balanced Random Forest model, and chose 2 as the number of trees. These parameters were selected after several hyperparameter tuning on multiple tiles to maximize the model performance. 

Finally, for each tile, the trained model was used to generate a LC label at 10m resolution that was then passed to annotators for validation.

\end{document}